\documentclass{article}

    \PassOptionsToPackage{numbers, compress}{natbib}

     \usepackage[final]{neurips_2019}

\usepackage[utf8]{inputenc} 
\usepackage[T1]{fontenc}    
\usepackage{hyperref}       
\usepackage{url}            
\usepackage{booktabs}       
\usepackage{amsfonts}       
\usepackage{nicefrac}       
\usepackage{microtype}      

\usepackage{amsmath}
\usepackage[ruled,vlined]{algorithm2e}
\usepackage{graphicx}
\graphicspath{ {./images/} }

\title{Mask CycleGAN : Unpaired Multi-modal Domain Translation with Interpretable Latent Variable}

\author{%
  Minfa Wang \\
  Stanford University\\
  \texttt{minfa@stanford.edu} \\
}

\begin{document}

\maketitle

\begin{abstract}
We propose Mask CycleGAN, a novel architecture for unpaired image domain translation built based on CycleGAN \cite{cgan}, with an aim to address two issues: 1) uni-modality in image translation and 2) lack of interpretability of latent variables. Our innovation in the technical approach is comprised of three key components: masking scheme, generator and objective. Experimental results demonstrate that this architecture is capable of bringing variations to generated images in a controllable manner and is reasonably robust to different masks.
\end{abstract}

\section{Introduction}

CycleGAN \cite{cgan} is a popular approach for unpaired image-to-image translation between two domains. It has been proven to be effective in a wide variety of domain translation tasks, including horse-zebra, apple-orange, summer-winter, etc. While it keeps inspiring generative modelling community to build up more and more applications and research ideas, CycleGAN has its limitations too. One notable limitation is that the translation is deterministic and hence lack of variation. People have discovered that achieving multi-modality through CycleGAN is challenging \cite{aug_cgan, bcgan}, largely due to the reason that the source image is at high dimensionality and usually causes the generator network to ignore other noise sampled from other distribution. A natural idea to enable multi-modal image generation is by introducing additional latent variables, which are often modeled as Gaussian distributions. However, samples from Gaussian distributions are generally lack of interpretability.

\begin{figure}[h]
  \centering
  \includegraphics[width=\textwidth]{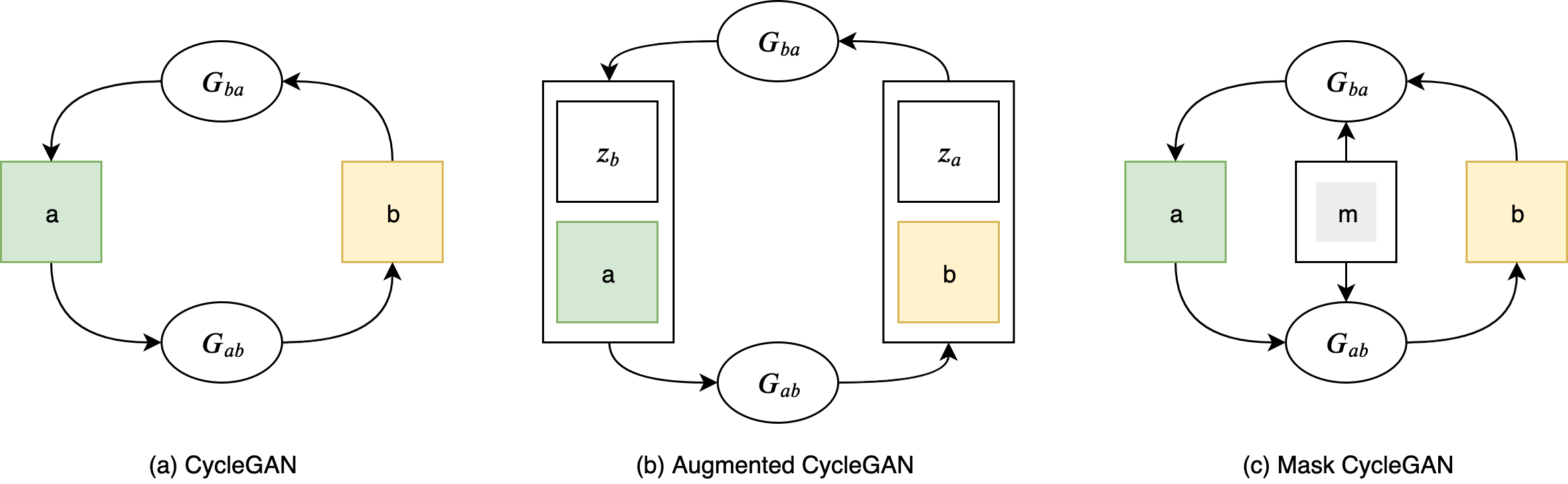}
  \caption{Comparison of variants of CycleGAN.}
  \label{architecture}
\end{figure}

\textbf{Mask CycleGAN} aims to address both issues above by using pixel mask as latent variables. Figure \ref{architecture} shows a high-level overview of our architecture and comparison with other popular architectures. In later sections, we will elaborate the architecture in details. We will show its formulation is a full generalization of CycleGAN, and hence is at least equally expressive. Moreover, the pixel mask offers great control of the image generation outcome.

\section{Related Work}

\paragraph{CycleGAN}\cite{cgan} is a popular approach for unpaired image-to-image translation between two domains. One of the most innovative ideas from this work is \textit{cycle consistency}, which encourages the mapping between two domains to be invertible, and indirectly alleviate the problem of mode collapse. CycleGAN is capable of generating visually appealing images.

\paragraph{Augmented CycleGAN} \cite{aug_cgan} brings multi-modality into CycleGAN by augmenting it with two latent variables $Z_a$ and $Z_b$, and corresponding encoders $E_a$, $E_b$ and discriminators $D_{Z_a}$, $D_{Z_b}$. The latent variables are optimized through the minimax game between the encoder and the discriminator.

\paragraph{BicycleGAN} \cite{bcgan} is an architecture for paired multi-modal image-to-image translation. The name Bicycle refers to the two cycles from 1) Conditional Variational Autoencoder GAN (\textit{cVAE-GAN}) and 2) Conditional Latent Regressor GAN (\textit{cLR-GAN}).

\paragraph{Image inpainting} is a technique to reconstruct a image patch from a partially covered or blurred image. The input of image inpainting can be thought as a clear source image applied on a mask, which is similar to our setup. Some of the works \cite{inpaint} in this field inspires the generator design of the Mask CycleGAN to impose soft constraint on pixel invariance on certain image region during transformation.

\paragraph{Attention} is technique to discover and make use of region of interests from the input by assigning different weights to different parts of the input. When the input is image, the weight is typically computed at pixel level, and the attention weight map can be thought of as a soft mask. People \cite{attention} have attempted to use unsupervised attention mask to improve generative modelling.

\section{Problem statement}

The task for CycleGAN-alike architectures is unpaired image-translation between two domains. At inference time, an image from source domain will be given as input, and the output will be an image from target domain, retaining the basic features of the input image. The mathematical formulation is introduced in the Notation section below.

\subsection{Notation}

We introduce the following terminologies to help elaborate the technical approach. Same as CycleGAN, we have $a$ as a sample image from the source domain, $b$ as a sample from the destination domain, and $G_{AB}$ as generator mapping an image from source to destination. With mask $m$, we could derive the following properties: $\tilde{b} = G_{AB}(a, m)$ as the fake image in destination domain, $a' = G_{BA}(\tilde{b}, m)$ as the recovered image in source domain. Other terms like $\tilde{a}$, $b'$ and $G_{BA}$ are defined symmetrically.

Regarding the mask, we call \textit{masked region} to be the region where the information of the pixel values of the image is kept, and the \textit{contextual region} to be the region outside the \textit{masked region}.

\section{Approach}

Our architecture is based on CycleGAN. Please refer to \cite{cgan} for more details of its original architecture. The sections below elaborate the key modifications we made to incorporate masks into the whole system.

\subsection{Masking}

\subsubsection{Discussion: binary vs. continuous mask}
\paragraph{Binary mask} a mask that has value 1 on its masked area, and 0 elsewhere. It is a kind of \textbf{hard} mask, eliminating all the information from the contextual region of the image.

\paragraph{Continuous mask} a mask that has float number between 0 and 1 for all its dimensions, representing the weights of different pixels of the image. It is a \textbf{soft} mask where the boundary between masked and contextual regions could be blurry, and the information for contextual region will be partially retained after the mask is applied on the image.

One of the reasons that it is challenging to introduce multi-modality in CycleGAN is that when you feed a concatenation of image features and latent variables as the input to the generator, the generator quickly learns that the dimensions of the latent variables provides little additional values in optimizing the overall objective, and hence zeros out those dimensions, forfeiting multi-modality.

Thanks to the interpretability of the mask, we can re-design the interaction between the mask and the input image to force the generator to respect the effect of the mask. The revamped generator design is detailed in the following Generator section. To avoid unexpected information leaking to the generator, we choose binary masks in this work. Below is a list of masking schemes that we have considered.

\subsubsection{Centered-square masking scheme}

The masked region always is centered in the image, has square shape, and has a size of $(0.5 | 0.8 | 1.0)$ * image size. Figure \ref{masks}(a) provides an visualization of the masks generated from this scheme.

The centered-square masking scheme is simple to understand and fast to evaluate. On the other hand, it is very limited on the amount variation it is capable of producing.

\subsubsection{Multi-rectangles masking scheme}

Multi-rectangles masking scheme, elaborated in Algorithm \ref{alg:multi_rectangles}, is a generalization of the centered-square masking scheme. It allows more variations in size, position and (compound) shape of the mask, encouraging the generator to generalize better. The limitation of this masking scheme is that it still produces rectangular edges. Figure \ref{masks}(b) provides an visualization of some samples generated from this masking scheme.

\begin{algorithm}[H]
\caption{Multi-rectangles masking scheme}
\label{alg:multi_rectangles}
\SetAlgoLined

\tcp{Minimal number of rectangles to draw.}
Set MinMaxNumRects = 5, MinNumRects = uniform(1, MinMaxNumRects) \;

\tcp{Minimal accumulative area relative to the whole image area.}
Set MinSumRelArea = 0.15 \;

\tcp{Size of image; lower and upper bound of individual rectangle size.}
Set Size = 128, MinRectSize = Size / 10, MaxRectSize = Size \;

\tcp{Initialize variables to keep track of status in the loop.}
Set numRects = 0 \;
Set sumRelArea = 0.0 \;
Set mask = zeros(3, Size, Size)  \;
 
 \While{numRects < MinNumRects \textbf{or} sumRelArea < MinSumRelArea}{
      \tcp{Randomly generate the top left corner of the rectangle.}
      i0 = uniform(0, Size - MinRectSize) \;
      j0 = uniform(0, Size - MinRectSize) \;
    
      \tcp{Randomly generate the bottom right corner of the rectangle.}
      i1 = uniform(i0 + MinRectSize, min(i0 + MaxRectSize, Size)) \;
      j1 = uniform(j0 + MinRectSize, min(j0 + MaxRectSize, Size)) \;
    
      \tcp{Draw rectangle on the mask. Update intermediate states.}
      mask[:, i0:i1, j0:j1] = 1 \;
      numRects += 1 \;
      sumRelArea += ((i1 - i0) * (j1 - j0)) / (Size * Size) \;
 }
\end{algorithm}

Some properties of this masking scheme:

\begin{itemize}
    \item When MinRectSize = 1 and MaxRectSize = 2, this masking scheme is equivalent to sample individual pixels independently.
    \item When MinSumRelArea = 1, this masking scheme will always generate the full mask, making the whole algorithm become equivalent to CycleGAN.
    \item The time complexity of this algorithm is O(max(MinNumRects, MinRelArea / (MinRectSize / Size)$^2$)).
\end{itemize}

\subsubsection{Binary-attention masking scheme}

If we have access to some pre-trained image attention network, which is able to produce a pixel level attention map, then we could potentially convert it to a binary mask by binarizing the attention weights on all pixels based on some threshold.

\begin{figure}
  \centering
  \includegraphics[width=\textwidth]{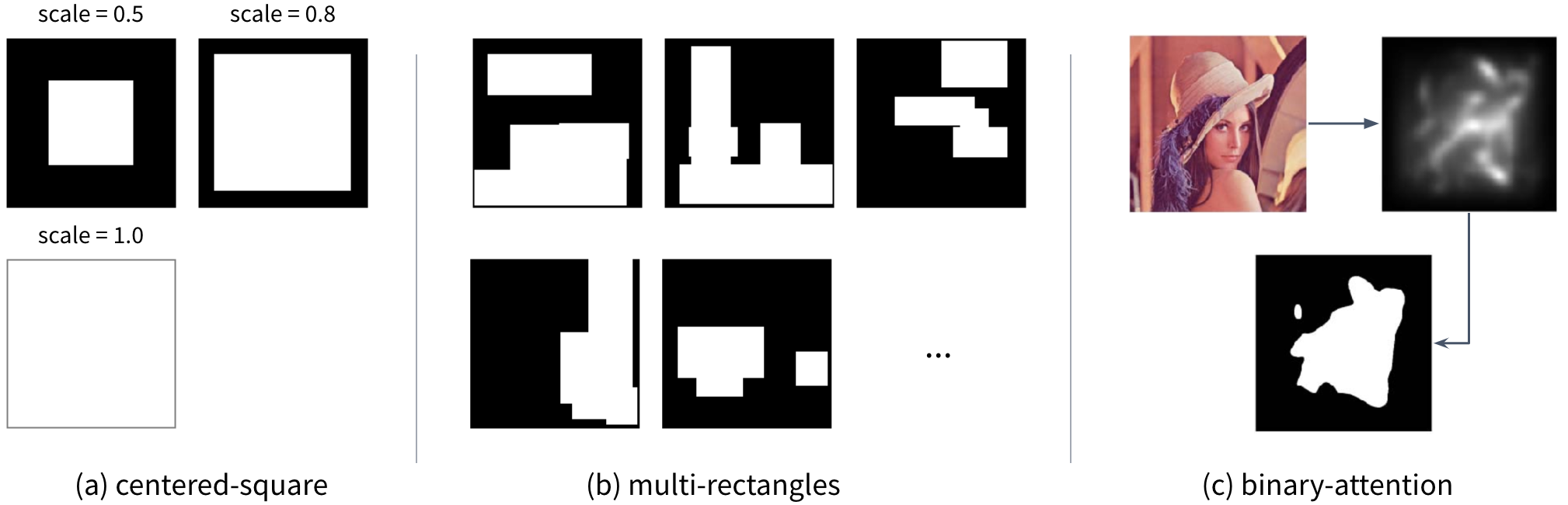}
  \caption{Visualization and comparison of different Masking schemes. (c) is adopted from an image in \cite{attention_image}.}
  \label{masks}
\end{figure}

Figure \ref{masks}(c) provides an illustration of the binary-attention masking scheme.

\subsection{Generator}

\begin{figure}[h]
  \centering
  \includegraphics[width=\textwidth]{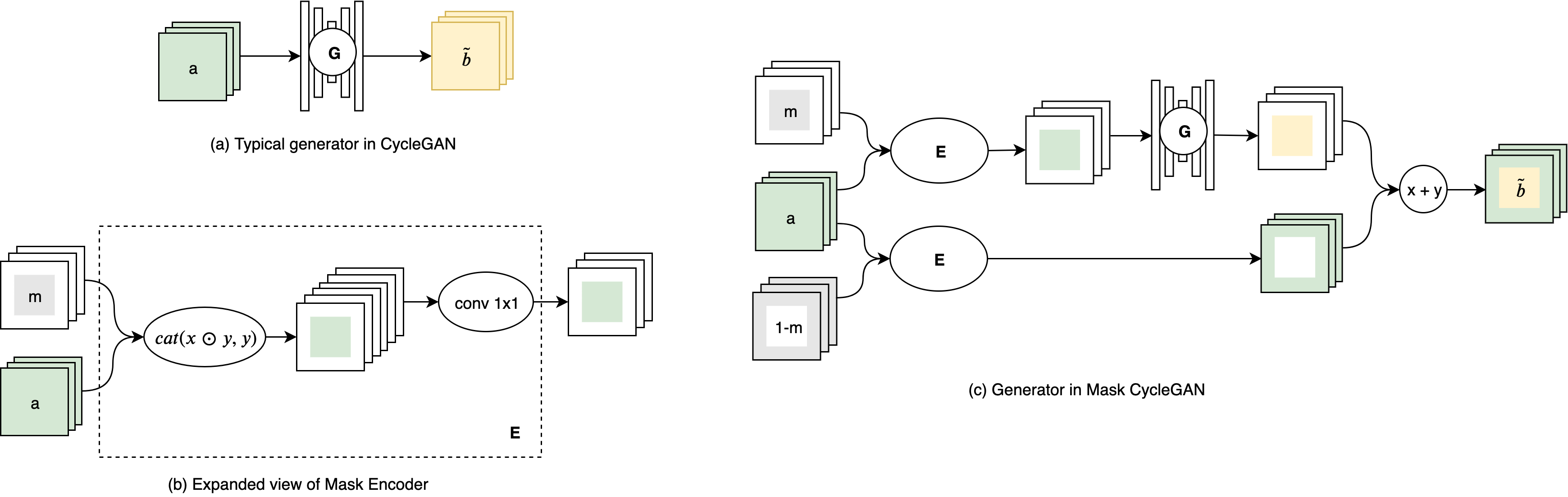}
  \caption{Comparison of generators. (b) is the expanded view of the mask encoder $E$ used in (c).}
  \label{generator}
\end{figure}

The generator design is one of the critical parts of the architecture. Figure \ref{generator}(a) shows the plain-vanilla generator used in original CycleGAN that maps a source image to a destination image. It is easy to see that the mapping is deterministic with a fixed source image. In literature, when people try to introduce the additional latent variable $z$, a common approach \cite{bcgan, aug_cgan} is concatenates the latent variable vector to some intermediate vector representation of the source image. The drawbacks of this approach are 1) the influence of latent variable is hard to interpret and control, and 2) empirically the generator often tends to ignore variations in the latent variable.

Since our latent variable, $m$, is interpretable, we could design its interaction with the source image in a way that enforces the masking behavior. The interaction is defined through the \textbf{mask encoder} $E$, with its architecture detailed in Figure \ref{generator}(b). The full generator design is shown in Figure \ref{generator}(c), where complicated domain translation logic, captured in $G$, is required to only depend on masked region of the source image.

\subsection{Losses}

Similar to CycleGAN, we optimize the model by minimizing three pairs of losses. We have made modifications on each pair to accommodate the architecture change introduced by the mask.

\paragraph{GAN loss} For each triplet $(a, \tilde{a}, m$), we will have two discriminators $D_{AF}$ and $D_{AM}$. The first generator $D_{AF}$ is functionally equivalent to the discriminator $D_A$ in original CycleGAN, which is trying to distinguish a true image from a generated one for the source distribution, and implicitly encourages the generator to produce overall coherent image inside and outside the masked area. The second discriminator $D_{AM}$ is responsible for the same task for the masked image pair. We have the final GAN loss as the normalized weighted sum of the two discriminators' losses. Formally:
$$
\begin{aligned}
\mathcal{L}_{GAN}^{AF} &= \mathbb{E}_{a \sim A}[ \log D_{AF}(a) ] + \mathbb{E}_{\tilde{a} \sim \tilde{A}}[ \log (1 - D_{AF}(\tilde{a})) ]
\\
\mathcal{L}_{GAN}^{AM} &= \mathbb{E}_{a \sim A}[ \log D_{AM}(a \odot m) ] + \mathbb{E}_{\tilde{a} \sim \tilde{A}}[ \log (1 - D_{AM}(\tilde{a} \odot m)) ]
\\
\mathcal{L}_{GAN}^{A} &= \lambda_{GAN}^{M} \mathcal{L}_{GAN}^{AM} + (1 - \lambda_{GAN}^{M}) \mathcal{L}_{GAN}^{AF}
\end{aligned}
$$
There are two remarks. First, we need to introduce the second discriminator $D_{AM}$ because despite the fact that generator tries to produce coherent images, there would generally exists some discontinuity around the mask boundary, making the task relatively easy for $D_A$, and in turn causes gradient vanishing problems to the generator. Even if the generator manages to learn, without $D_{AM}$, the generator would tend to ignore the mask and suffer from mode collapse. Second, when $\lambda_{GAN}^{M} L_{GAN}^{AM} = 0$, this objective falls back to the GAN objective in original CycleGAN.

\paragraph{Cycle loss} facilitates cycle-consistency for translation between two domains. The high-level idea is that when a image goes through a forward mapping and then a backward mapping, it should recover to itself, which requires the forward mapping to preserve the information of the source image. This is an effective method to prevent mode collapse. Our loss objective is the normalized weighted sum of the cycle losses for the masked and context areas.
$$
\begin{aligned}
\mathcal{L}_{CYC}^{A} = \lambda_{CYC}^{M} || (a - a') \odot m ||_1 + (1 - \lambda_{CYC}^{M}) || (a - a') \odot (1 - m) ||_1
\end{aligned}
$$
Different weights are applied for masked and context areas, and the weights could be adjusted to control how strict we hope the generator to keep the pixels in the context area intact. When $\lambda_{CYC}^M = 0.5$, this loss falls back to the cycle loss in original CycleGAN (with a scale factor).

\paragraph{Identity loss} states that when you try to map an image from the destination domain to the destination domain, it should map to itself. Formally:
$$
\begin{aligned}
\mathcal{L}_{IDT}^{A} = || a - G_{BA}(a, m) ||_1
\end{aligned}
$$
Note this formulation is identical to the identity loss in original CycleGAN. We don't need special treatment for the masked area because we intentionally want this specific mapping to be invariant of the mask.

\subsection{Full objective}

The full objective is
$$
\mathcal{L} = (\mathcal{L}_{GAN}^{A} + \mathcal{L}_{GAN}^{B}) + \lambda_{CYC} (\mathcal{L}_{CYC}^{A} + \mathcal{L}_{CYC}^{B}) + \lambda_{IDT} (\mathcal{L}_{IDT}^{A} + \mathcal{L}_{IDT}^{B})
$$

\subsection{Algorithm}

\begin{algorithm}[H]
\SetAlgoLined
 initialize weights and optimizers \;
 initialize $G_{AB}$, $G_{BA}$, $D_{AF}$, $D_{BF}$, $D_{AM}$, $D_{BM}$ \;
 \While{True}{
  Read next batch $a$, $b$ \;
  Sample $m$ from $M$ \;
  $\tilde{b}$ = $G_{AB}(a, m)$, $\tilde{a}$ = $G_{BA}(b, m)$ \;
  $a'$ = $G_{BA}(\tilde{b}, m)$, $b'$ = $G_{AB}(\tilde{a}, m)$ \;

  \tcp{--- optimize generator below ---}
  l\_cyc\_a = $\lambda_{CYC}^{M}$ * cyc\_loss(a * m, $a'$ * m) + (1 - $\lambda_{CYC}^{M}$) * cyc\_loss(a * (1 - m), $a'$ * (1 - m)) \;
  l\_cyc\_b = $\lambda_{CYC}^{M}$ * cyc\_loss(b * m, $b'$ * m) + (1 - $\lambda_{CYC}^{M}$) * cyc\_loss(b * (1 - m), $b'$ * (1 - m)) \;
  
  l\_idt = idt\_loss(a, $G_{BA}$(a, m)) + idt\_loss(b, $G_{AB}$(b, m)) \;
  
  l\_gan\_a = (1 - $\lambda_{GAN}^{M}$) * gan\_loss($D_{AF}$($\tilde{a}$)) +
              $\lambda_{GAN}^{M}$ * gan\_loss($D_{AM}$($\tilde{a}$ * m))) \;
  l\_gan\_b = (1 - $\lambda_{GAN}^{M}$) * gan\_loss($D_{BF}$($\tilde{b}$)) +
              $\lambda_{GAN}^{M}$ * gan\_loss($D_{BM}$($\tilde{b}$ * m))) \;
              
  l\_g = (l\_gan\_a + l\_gan\_b) + $\lambda_{CYC}$ * (l\_cyc\_a + l\_cyc\_b) + $\lambda_{IDT}$ * l\_idt \;
  l\_g.backward() \;
  optimizer\_g.step(l\_g) \;
  
  \tcp{--- optimize discriminator below ---}
  l\_da = (1 - $\lambda_{GAN}^{M}$) * d\_loss($D_{AF}$, a, $\tilde{a}$) + $\lambda_{GAN}^{M}$ * d\_loss($D_{AM}$, a * m, $\tilde{a}$ * m) \;
  l\_db = (1 - $\lambda_{GAN}^{M}$) * d\_loss($D_{BF}$, b, $\tilde{b}$) + $\lambda_{GAN}^{M}$ * d\_loss($D_{BM}$, b * m, $\tilde{b}$ * m) \;
  l\_d = l\_da + l\_db \;
  l\_d.backward() \;
  optimizer\_d.step(l\_d) \;
 }
 \caption{Masked CycleGAN}
\end{algorithm}

\section{Experiments}

\subsection{Setup}

We evaluated the model on the following datasets: \textit{MNIST-SVHN}, \textit{Horse-Zebra}, \textit{Monet-Photo}, \textit{Vangogh-Photo}. The image resolution for all datasets is 128x128, except for MNIST-SVHN, which is 32x32. For all the experiments, we used the same hyper-parameter settings as follow: $\lambda_{GAN}^{M} = 0.7$, $\lambda_{CYC}^{M} = 0.3$, $\lambda_{CYC} = 10$ and $\lambda_{IDT} = 5$.

We experimented with the centered-square and multi-rectangles masking schemes, and evaluated the algorithm both qualitatively and quantitatively. For quantitative analysis, we used the FID score of test set and the set generated with full mask as \textbf{baseline}.

\subsection{Quantitative Results}

\paragraph{Frechet Inception Distance (FID)} \cite{fid} is a method to measure the performance of a generative model by computing the following distance on the inception feature representation of the generated data distribution and true data distribution fitted by multivariate Gaussian. We use FID as the main quantitative metric for our model.

\subsubsection{Train with centered-square masking scheme}

\begin{figure}
  \centering
  \includegraphics[width=\textwidth]{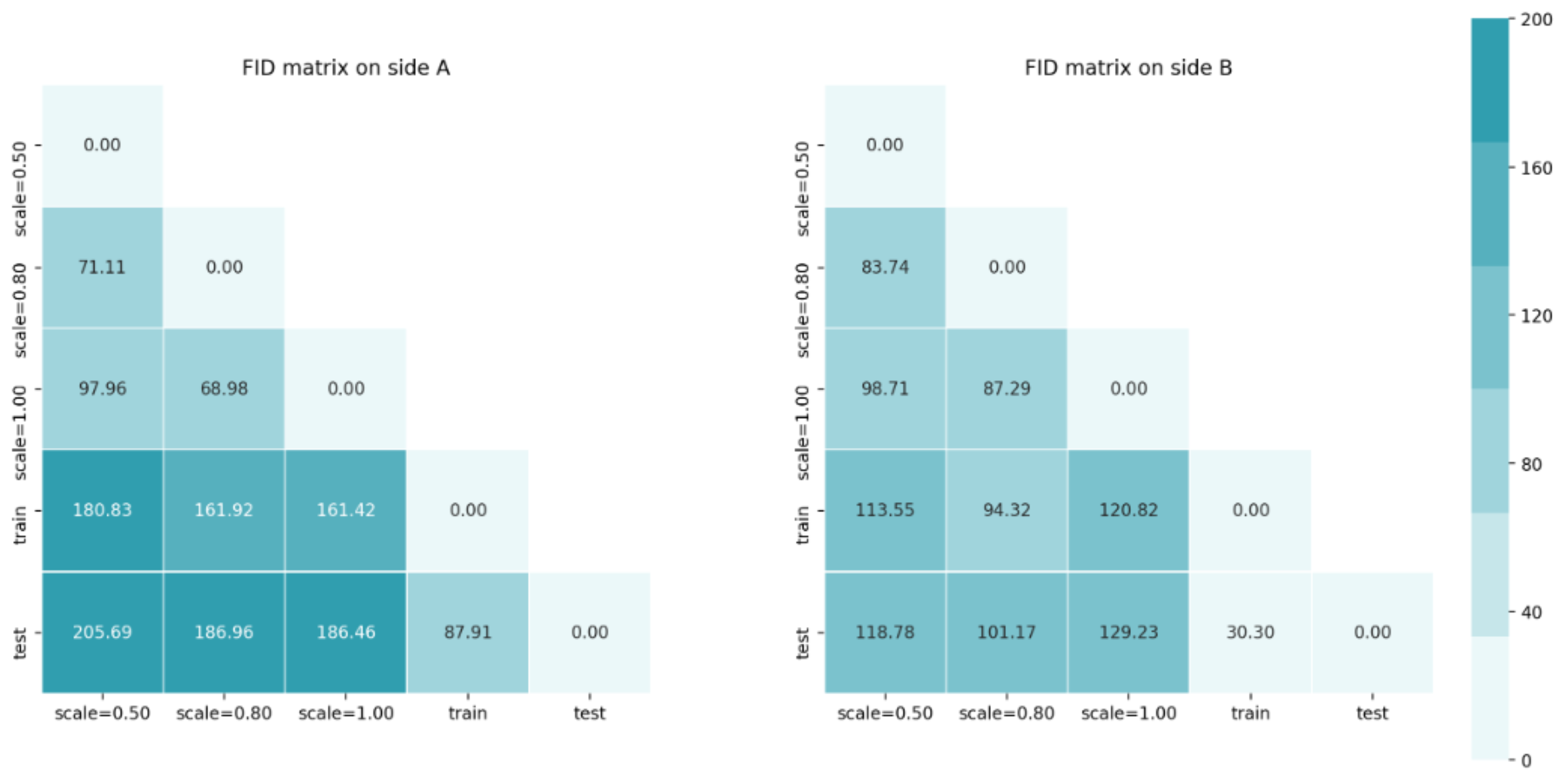}
  \caption{FID matrix of Mask CycleGAN trained with \textbf{centered-square} masking scheme on Horse-Zebra dataset. (a) shows FID scores among real and generated A:horse datasets. (b) shows FID scores among real and generated B:Zebra datasets.}
  \label{fid_matrix}
\end{figure}

We computed FID scores for various pairs of datasets derived from the \textit{Horse-Zebra} dataset. The results are represented in two matrices shown in Figure \ref{fid_matrix}, where (a) shows FID scores for various horse datasets, and (b) shows FID scores for various zebra datasets.

There are several conclusions we could draw from the matrices:

\begin{itemize}
    \item $FID_*(scale=*, train) < FID_*(scale=*, test)$. The generator was trained on the training dataset, and hence would be able to emulate the training data distribution better.
    \item $FID_A(train, test) > FID_B(train, test)$. According to FID, the horse dataset has more variations in style. This characteristic matches with our human-eye judgement.
    \item $FID_A(scale=*, test)$ > $FID_B(scale=*, test)$. The generator performs worse in emulating the horse distribution. This could be attributed to the intrinsic difficulty of horse dataset.
    \item $\boldsymbol{FID_B(scale=0.8, test) < FID_B(scale=1.0, test)}$. This is probably the most interesting finding. Usually, the larger the mask scale, the more information is exposed to the generator, and hence generator has more expressive power to fit a distribution. However, for the horse dataset, we see that the model performs better with the comparably smaller mask. One explanation is that typically the object of interest is presented around the center of the image, and a smaller mask actually filters out some noise and produces some mild regularization.
\end{itemize}

\subsubsection{Train with multi-rectangles masking scheme}

\begin{figure}
  \centering
  \includegraphics[width=\textwidth]{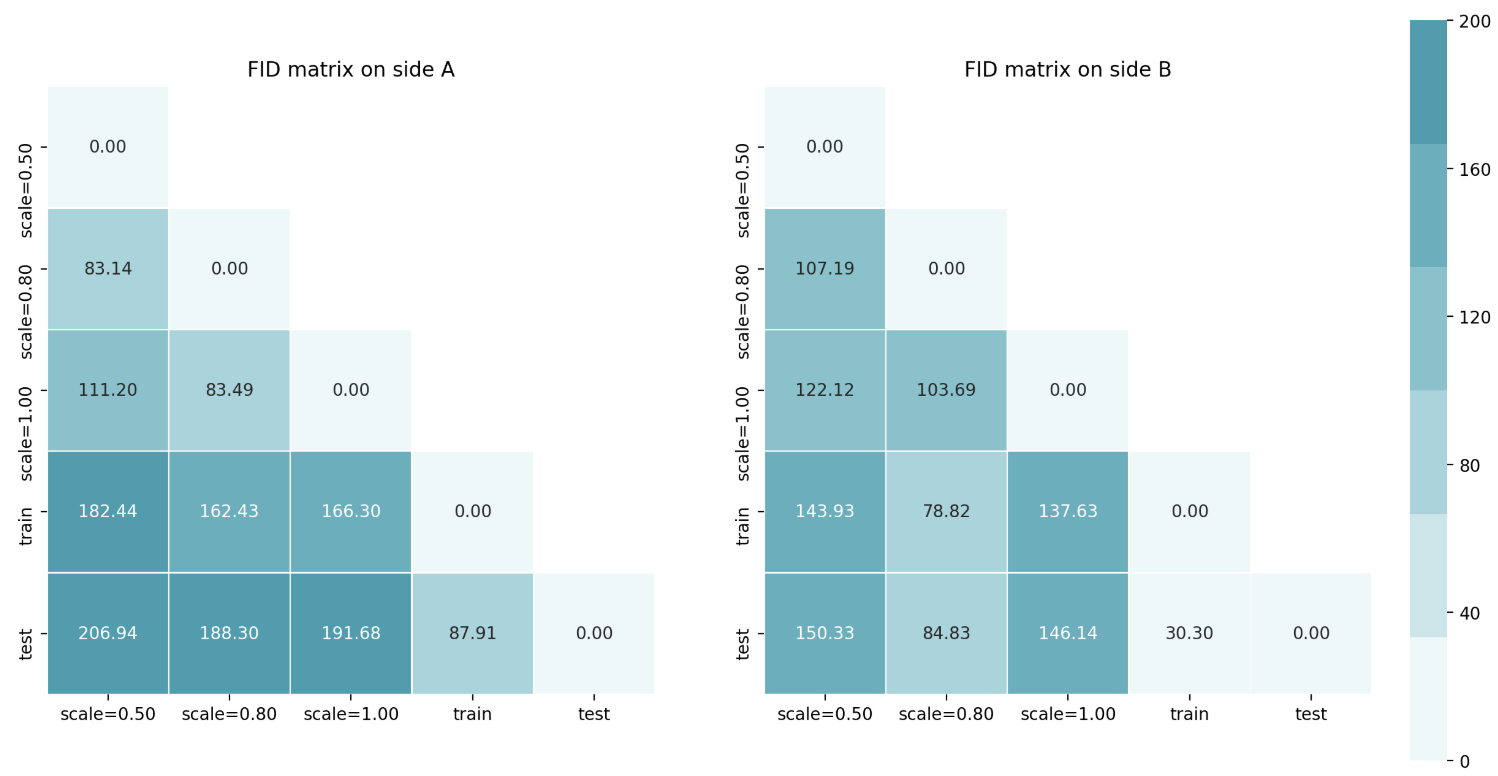}
  \caption{FID matrix for model with \textbf{multi-rectangles} masking scheme, evaluated on horse-zebra dataset.}
  \label{fid_matrix_multi_rects}
\end{figure}

We also conducted FID evaluation on the model with multi-rectangles masking scheme. The result is shown in Figure \ref{fid_matrix_multi_rects}.

From the matrices, we could draw the same conclusions as the previous experiment. In addition, what is worth noting is that regardless of masking schemes in training, we used centered-square masking scheme in inference. Nonetheless, on $FID_B(scale=0.8, test)$, the model trained with multi-rectangles masking scheme \textbf{outperforms} the model trained with centered-square masking scheme: 84.83 vs. 101.17. This is a promising sign that more variations in masks in training could help the model generalize better in inference.

\subsection{Qualitative Results}

\begin{figure}
  \centering
  \includegraphics[width=\textwidth]{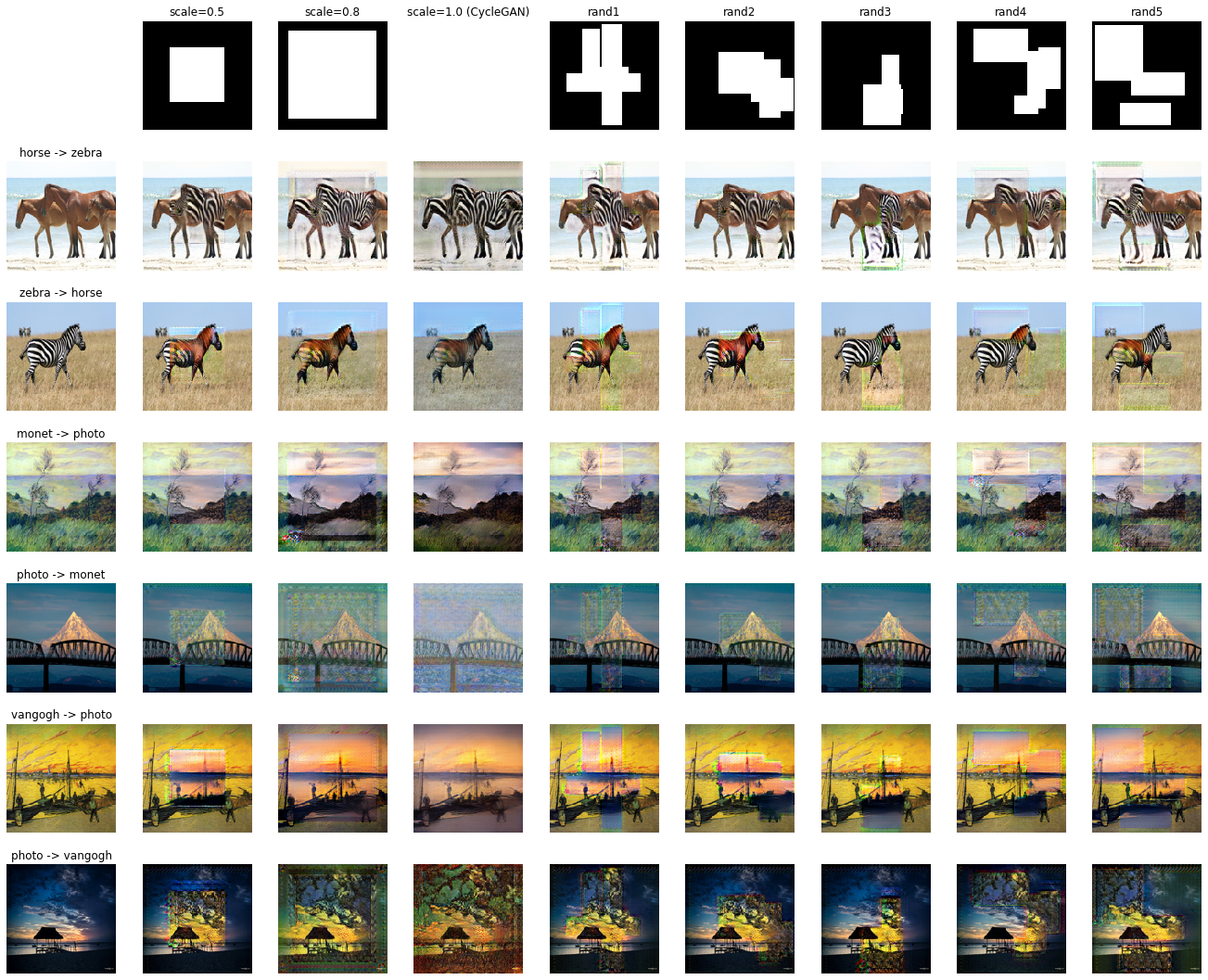}
  \caption{Outputs grid of Mask CycleGAN evaluated on several datasets with different masks. First row shows different masks used on different columns. First column shows the source images and the related tasks. Other cells show the generator output with the corresponding source image and mask as inputs.}
  \label{output_grid}
\end{figure}

We examine the qualitative results of the model through a output grid shown in Figure \ref{output_grid}. The model used in the grid was trained only on centered-square masks, but was evaluated on both centered-square and multi-rectangles masks. It is shown from the output that Mask CycleGAN is robust in translation across many image domains, and is able to generalize to work with mask it has never seen during training.

\subsubsection{Generalization to round mask}

\begin{figure}[h]
  \centering
  \includegraphics[width=\textwidth]{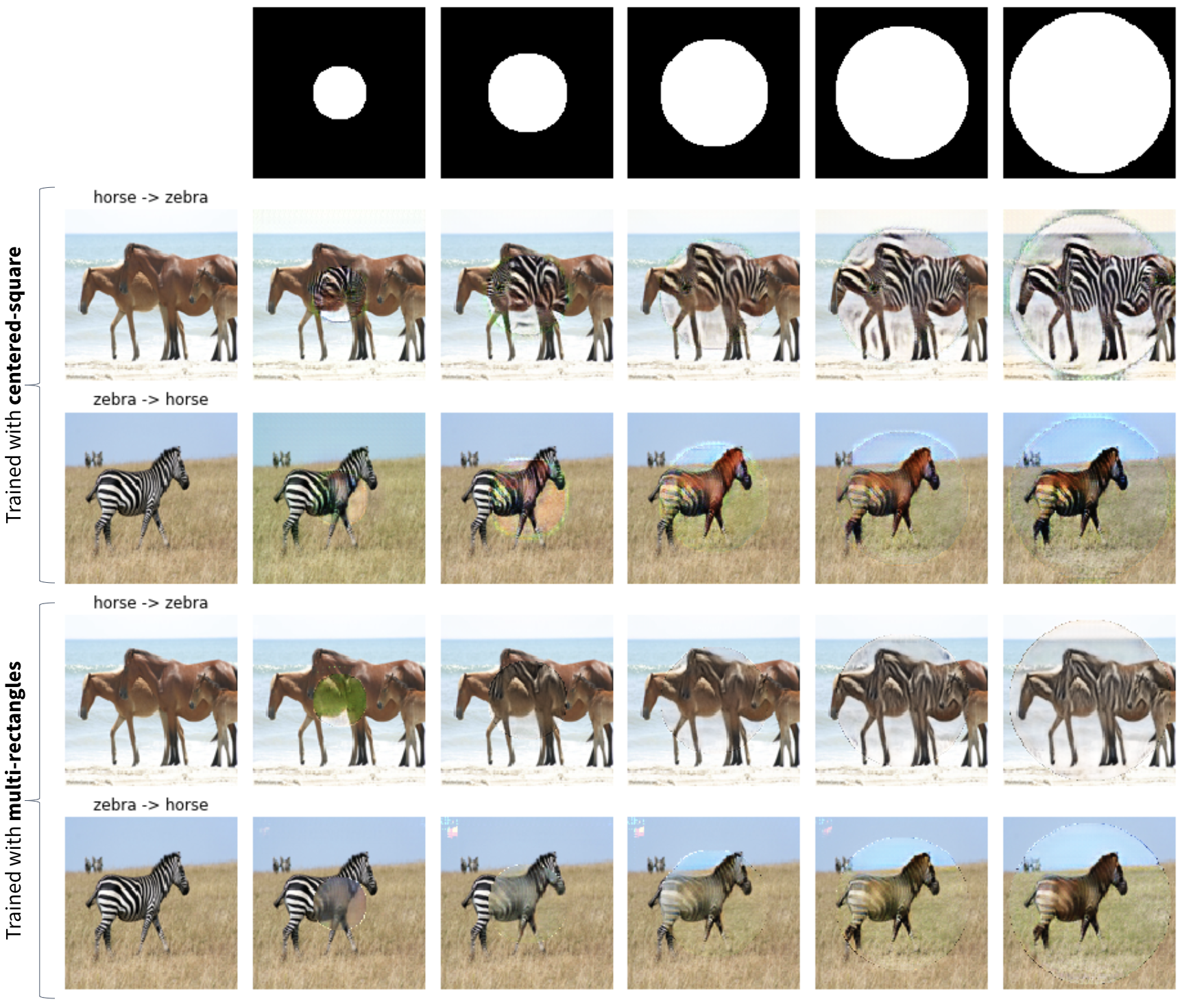}
  \caption{Models evaluated on round masks.}
  \label{output_grid_round}
\end{figure}

One interesting question is that if only trained with rectangular masks, will the model be able to generalize to round masks? We did the analysis by feeding the generator with round masks of different scales and the outputs are shown in Figure \ref{output_grid_round}. Interestingly, model trained with centered-square masking scheme appears to generalize better to round masks.

\section{Conclusion}

In this work, we proposed a novel generative modelling algorithm called \textbf{Mask CycleGAN}. We introduced the motivation of the idea, the mathematical formulation, and quantitative and qualitative evaluations of the algorithm. We illustrate that Mask CycleGAN is capable of bringing variations in CycleGAN-alike generators in a controllable manner. We believe this architecture will open doors for a series of interesting applications.

In the future, we plan to further improve the robustness of the algorithm along two directions: 1) design of generator, where we may experiment with more sophisticated architecture applied on the context region of the image; 2) explore different masking schemes like binary-attention to help the model generalize better to more variations of masks during inference.

\section{Acknowledgement}

Special thanks to Kristy Choi for her feedback and advice offered in the development of the project, and Sonya Chen for proofreading.

The code for this project is available at \url{https://www.github.com/minfawang/mask-cgan}.

\medskip
\bibliography{neurips_2019}

\begin{thebibliography}{7}
\providecommand{\natexlab}[1]{#1}
\providecommand{\url}[1]{\texttt{#1}}
\expandafter\ifx\csname urlstyle\endcsname\relax
  \providecommand{\doi}[1]{doi: #1}\else
  \providecommand{\doi}{doi: \begingroup \urlstyle{rm}\Url}\fi

\bibitem[Agarwal et~al.(2013)Agarwal, Bose, Maiti, Islam, and
  Sarkar]{attention_image}
C.~Agarwal, A.~Bose, S.~Maiti, N.~Islam, and S.~Sarkar.
\newblock Enhanced data hiding method using dwt based on saliency model.
\newblock pages 1--6, 09 2013.
\newblock ISBN 978-1-4673-6188-0.
\newblock \doi{10.1109/ISPCC.2013.6663414}.

\bibitem[Almahairi et~al.(2018)Almahairi, Rajeswar, Sordoni, Bachman, and
  Courville]{aug_cgan}
A.~Almahairi, S.~Rajeswar, A.~Sordoni, P.~Bachman, and A.~C. Courville.
\newblock Augmented cyclegan: Learning many-to-many mappings from unpaired
  data.
\newblock \emph{CoRR}, abs/1802.10151, 2018.
\newblock URL \url{http://arxiv.org/abs/1802.10151}.

\bibitem[Heusel et~al.(2017)Heusel, Ramsauer, Unterthiner, Nessler, Klambauer,
  and Hochreiter]{fid}
M.~Heusel, H.~Ramsauer, T.~Unterthiner, B.~Nessler, G.~Klambauer, and
  S.~Hochreiter.
\newblock Gans trained by a two time-scale update rule converge to a nash
  equilibrium.
\newblock \emph{CoRR}, abs/1706.08500, 2017.
\newblock URL \url{http://arxiv.org/abs/1706.08500}.

\bibitem[Mejjati et~al.(2018)Mejjati, Richardt, Tompkin, Cosker, and
  Kim]{attention}
Y.~A. Mejjati, C.~Richardt, J.~Tompkin, D.~Cosker, and K.~I. Kim.
\newblock Unsupervised attention-guided image to image translation.
\newblock \emph{CoRR}, abs/1806.02311, 2018.
\newblock URL \url{http://arxiv.org/abs/1806.02311}.

\bibitem[Pathak et~al.(2016)Pathak, Kr{\"{a}}henb{\"{u}}hl, Donahue, Darrell,
  and Efros]{inpaint}
D.~Pathak, P.~Kr{\"{a}}henb{\"{u}}hl, J.~Donahue, T.~Darrell, and A.~A. Efros.
\newblock Context encoders: Feature learning by inpainting.
\newblock \emph{CoRR}, abs/1604.07379, 2016.
\newblock URL \url{http://arxiv.org/abs/1604.07379}.

\bibitem[Zhu et~al.(2017{\natexlab{a}})Zhu, Park, Isola, and Efros]{cgan}
J.~Zhu, T.~Park, P.~Isola, and A.~A. Efros.
\newblock Unpaired image-to-image translation using cycle-consistent
  adversarial networks.
\newblock \emph{CoRR}, abs/1703.10593, 2017{\natexlab{a}}.
\newblock URL \url{http://arxiv.org/abs/1703.10593}.

\bibitem[Zhu et~al.(2017{\natexlab{b}})Zhu, Zhang, Pathak, Darrell, Efros,
  Wang, and Shechtman]{bcgan}
J.~Zhu, R.~Zhang, D.~Pathak, T.~Darrell, A.~A. Efros, O.~Wang, and
  E.~Shechtman.
\newblock Toward multimodal image-to-image translation.
\newblock \emph{CoRR}, abs/1711.11586, 2017{\natexlab{b}}.
\newblock URL \url{http://arxiv.org/abs/1711.11586}.

\end{thebibliography}

\newpage

\section{Appendix}

\begin{figure}[h!]
  \centering
  \includegraphics[width=\textwidth]{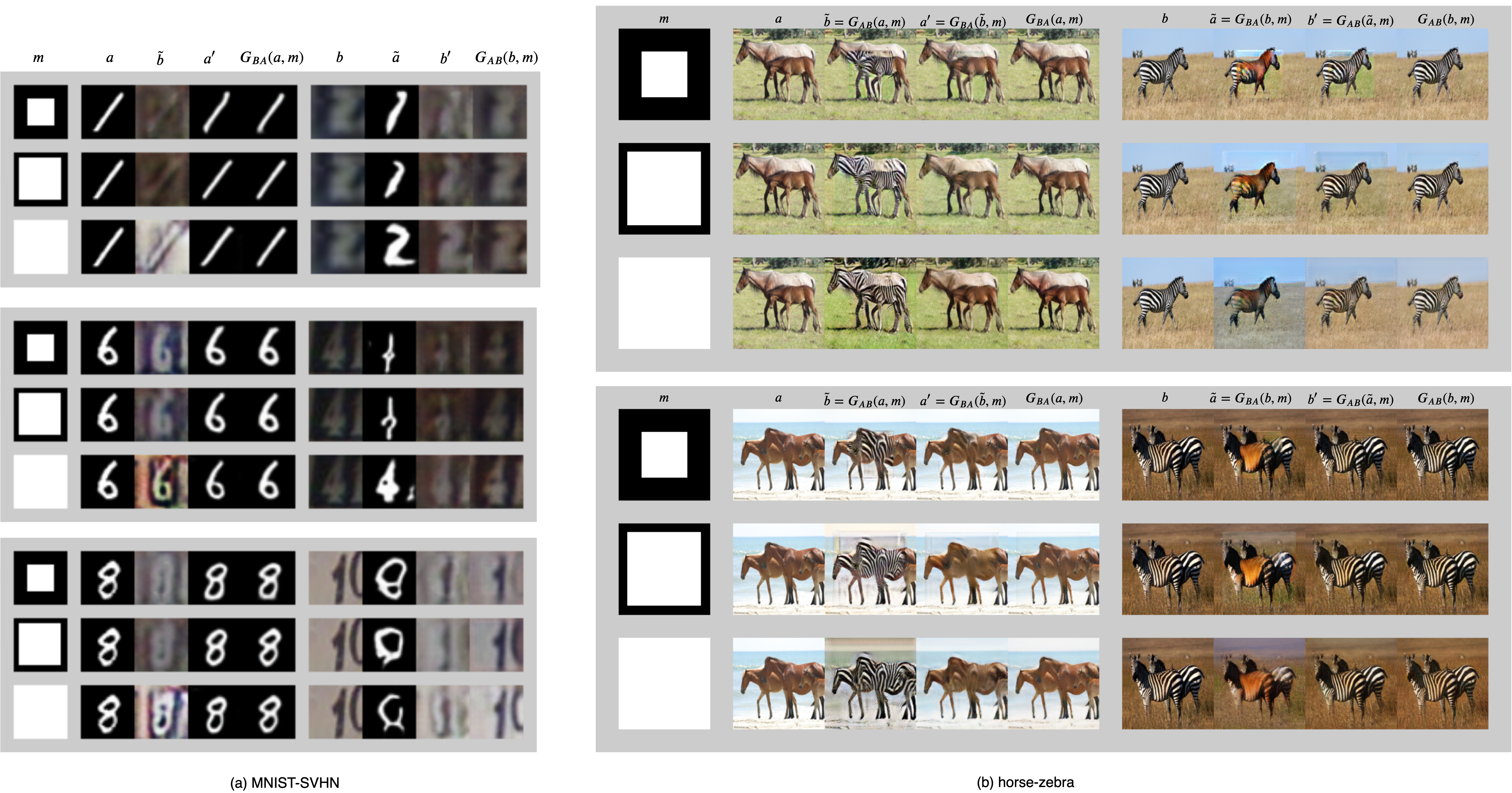}
  \caption{Outputs of Mask CycleGAN. (a) shows the results from MNIST-SVHN translation. (b) shows horse-zebra translation. Please zoom in to inspect the details.}
  \label{results}
\end{figure}

\end{document}